# Deep Learning Frameworks for Pavement Distress Classification: A Comparative Analysis

Vishal Mandal
Department of Civil & Environmental Engineering
University of Missouri Columbia
WSP USA
211 N Broadway, St. Louis MO 63108
vmghv@mail.missouri.edu

Abdul Rashid Mussah
Department of Civil & Environmental Engineering
University of Missouri Columbia
E2509 Lafferre Hall, Columbia, MO 65211
akm2fx@mail.missouri.edu

Yaw Adu-Gyamfi
Department of Civil & Environmental Engineering
University of Missouri Columbia
E2509 Lafferre Hall, Columbia, MO 65211
adugyamfiy@missouri.edu

*Abstract*— **Automatic detection and classification of pavement distresses is critical in timely maintaining and rehabilitating pavement surfaces. With the evolution of deep learning and high performance computing, the feasibility of vision-based pavement defect assessments has significantly improved. In this study, the authors deploy state-of-the-art deep learning algorithms based on different network backbones to detect and characterize pavement distresses. The influence of different backbone models such as CSPDarknet53, Hourglass-104 and EfficientNet were studied to evaluate their classification performance. The models were trained using 21,041 images captured across urban and rural streets of Japan, Czech Republic and India. Finally, the models were assessed based on their ability to predict and classify distresses, and tested using F1 score obtained from the statistical precision and recall values. The best performing model achieved an F1 score of 0.58 and 0.57 on two test datasets released by the IEEE Global Road Damage Detection Challenge. The source code including the trained models are made available at [1].**

*Keywords—deep convolutional neural networks, deep learning, pavement distress, road crack detection*

I. INTRODUCTION

The distresses in pavement surfaces present a potential threat to roadway and driving safety. For most state and local level transportation agencies, maintaining high quality of pavement surfaces is critical to their quotidian endeavor of keeping roadways safe and cornerstone for a sustainable transportation infrastructure. Detecting distresses timely is often regarded as one of the most crucial steps in limiting further degradation and maintaining high quality pavement surfaces. In 2019, the United States federal government spent a total of 29 billion dollars on infrastructure where almost half of the federal transportation spending went into highway and roadway infrastructure [2]. To optimize financial resources, there exists a need to periodically assess the condition of pavement surfaces and have its maintenance in check. These maintenance assessments could be either manual or have some form of automation applied.

In manual assessments, technicians determine the condition of pavements and give off a rating that is subject to vary depending on every other technician. Not to forget that these inspection techniques require domain expertise and field trips which can be tedious, unsafe and rather expensive [3]. To overcome these manual techniques, optimal and cost-effective procedures could be applied that automate the detection and characterization of pavement distresses. Recently, automated evaluation techniques utilizing vehicle mounted cameras have been used to detect and characterize pavement distresses [4, 5]. Similarly, some state highway agencies have been collecting pavement condition data through automatic 3D surveys that acquire high resolution images capable of identifying damaged pavement surfaces [6, 7]. In such images, the level of distresses can often be interpreted in the form of type, length and severity of pavement cracks. Since, cracking normally appears in the initial declining stages, proper detection techniques would allow for maintenance streams to take heed that can ultimately save both time and money. Also, as pavement surfaces only deteriorate 40 percent in quality during the first three quarters of their life-time, there exists a potential for restoration when detected and treated timely [8].

With aging infrastructure, state highway agencies have realized the importance of advanced computerized techniques in assessing pavement condition. By the end of 2012, over 35 state highway agencies in the US had deployed either automated or semi-automated image based techniques to obtain pavement cracking data [9]. In the current state of practice, image based computer vision techniques are quite common. Also, with the deteriorating highway quality in the US, where almost 1 out of every 5 miles of highway pavement warrants attention, the issue is certainly worth acting upon [10, 11]. The authors in this paper thus, lay out an integrated approach that builds up on some of the advanced deep learning frameworks, utilizing the recent advances in high performance computing to detect pavement distresses from images captured using a vehicle mounted smartphone.

The paper is structured systematically where the authors begin Section II discussing some of the related works relevant to this study, section III where an open-source pavement dataset is described, Section IV where the proposed method is discussed, Section V which broadly explains the experimental results and Section VI with final conclusions of the study.

## II. RELATED WORK

In the past few years, many computer vision based studies have emerged that focus on the automatic detection of pavement distresses. Some of these studies include the use of local binary patterns [12, 13], Gabor filters [14], shape based methods [15] and tree structure algorithms [16] among many others. Although these algorithms are generally superior, their inability at capturing discriminative features from images make them incapable at differentiating crack and non-crack image pixels. Also, these methods fail at accurately detecting distresses in real-world situations such as changing illumination and varied pavement textures. Deep learning, on the other hand has shown a tremendous potential at solving similar problems.

Recently, the application of deep learning in computer vision based detection of pavement distresses have been studied extensively. Based on how deep learning-based methods detect distresses, they can be broadly classified into three main categories viz. pure image classification, pixel-level segmentation and object detection based methods [17]. In pure image classification, an image is divided into overlapping blocks and then the block image is separated into classes. Afterwards, a deep convolutional neural network (DCNN) decides what image blocks contain distresses. Classification based methods utilizes both binary and multi-class differentiation techniques. Zhang et al. used convolutional neural networks (CNN) to reduce pavement images into smaller patch sizes and based on the output probability classified these patches as crack and no-crack [18]. Using a smart phone, modified GoogLeNet is used to organize image blocks in detecting pavement cracks [19]. Multi-class classification typically comes in handy when detecting different classes of cracks. In [20] Li et al. used DCNN to classify pavement cracks utilizing 3D images and further labeled those cracks into 5 different categories. In their approach, they deployed a total of 4 different CNNs for classification which despite having various receptive field sizes had little effect on the overall accuracy.

Similarly, in pixel-level segmentation, a score or label is assigned to every pixel value within an image. Fan et al. in [21] proposed a supervised DCNN to learn pavement crack textures from raw images by analyzing crack pixels within an image. To get to that, they used 4 convolutional, 2 max-pooling and 3 fully connected layers as their network architecture. Likewise, Jenkins et al. used a semantic segmentation approach founded on U-net for road crack detection [22]. Their U-net based approach acted as an encoder-decoder layered structure. In [23] Zou et al. proposed an end-to-end trainable DCNN to detect cracks. Their approach too uses an encoder-decoder based architecture that differentiates pavement image pixels into crack and non-crack background forms. Encoder-decoder functionalities are explained in [24] where multiple discriminative features are extracted from pavement crack images using only 4 convolutional and max pooling layers in encoder. Their method uses the multi-dilation module which can create inference from multi-featured cracks through the process of dilated convolution based on multiple rates. Furthermore, they optimized the multi-dilation features using the SE-Upsampling technique.

Likewise, object detection based methods typically locate distresses within a pavement image using bounding box approach. These methods first extract discriminative features from an image using a CNN which is followed by generating regions of interest (RoI) and finally detects objects through bounding box coordinates. YOLO [25] based crack detection is proposed in [26, 27] that can detect different classes of cracks in near real-time. Likewise, Li et al. in [28] deployed Faster R-CNN [29] to detect multiple crack types including potholes and alligator cracks. An important point to note is that Faster R-CNN is a two-stage detector where in its first stage, it generates RoI and further, passes down the region proposals for object classification and bounding box regression in its second stage. This approach of two-stage detection causes Faster R-CNN to obtain higher accuracy but the overall process remains slower compared to single-stage detector such as YOLO. In the current study, the authors have based their research on single-stage detectors that can detect pavement distresses in near real-time and can be scalable.

## III. DATA

Road images released by the IEEE Big Data, Global Road Damage Detection Challenge was used in this study [4, 30]. These images were taken from a vehicle mounted smartphone and captured scenes of urban and rural streets across three different countries namely, Japan, Czech Republic and India. The dataset comprised of altogether three sets: train, test1 and test2. The train set contained 10,506 images from Japan, 2,829 from Czech Republic and 7,706 from India. Images in the train set were pre-annotated end to end with their respective classes of pavement cracks which served as ground truths. Table I illustrates all four pavement cracks classified as longitudinal, transverse, alligator and pothole. Similarly, test1 and test2 contain images without annotations, and were used in evaluation of results. The total image count for test1 and test2 were 2,631 and 2,664 respectively.

TABLE I. Pavement Distresses with their Corresponding Classes

| Pavement Distress | Distress Class | Sample Image |
|---|---|---|
| Longitudinal Crack | D00 | 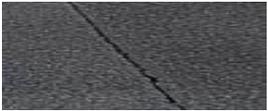 |
| Transverse Crack | D10 | 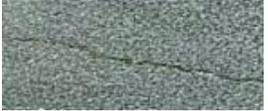 |
| Alligator Crack | D20 | 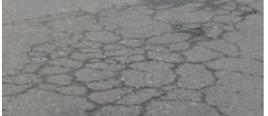 |
| Pothole | D40 | 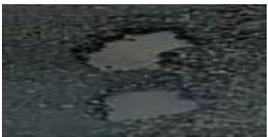 |

## IV. PROPOSED METHODOLOGY

The objective of this study is to develop a simplified pavement surface assessment system leveraging deep learning algorithms. The proposed framework should be capable of detecting distresses that would most likely differ from one another due to the variation in pavement textures among the following countries: Japan, Czech Republic and India. Also, each class of pavement distress is regarded as a distinguishable object and the proposed approach could be applied to any country based on fine-tuning of certain parameters. To learn the visual and textual patterns of each of these distress types, three single stage object detection algorithms, i.e. YOLO [31], CenterNet [32], and EfficientDet [33] were deployed. All these algorithms were implemented in Pytorch [34] and the network models were trained physically on an NVIDIA GTX 1080 Ti GPU and a cloud based Google Colab platform. Since, YOLO, CenterNet and EfficientDet detect objects using bounding boxes, the pavement distresses which were intrinsically different from each other need their visual patterns be analyzed within each of their bounding boxes. A brief description of these single stage object detectors is presented as follows.

### A. YOLO

You Only Look Once is a state-of-the-art object detection algorithm. For any object detector, there is a need to have a certain network size (i.e. resolution) to detect fine textures of pavement distresses, specific number of layers for an increased receptive field size and a significantly large number of parameters to further fine-tune. The most advanced iteration of YOLO achieves this requirement by using the CSPDarknet53 backbone containing 29 convolutional layers $3 \times 3$, receptive field of $725 \times 725$ and a total of 27.6 M parameters. Furthermore, the SPP block added over YOLO's CSPDarknet53 increases the size of receptive field without affecting its operating speed. Similarly, PANet is used for parameter aggregation from different levels of backbone. This also goes to different levels of detector. Some of the advanced features such as weighted-residual-connections, cross-stage-partial-connections, cross mini-batch, normalization (CmBN), and self-adversarial-training make YOLO extremely efficient. Also, to achieve higher precision values, the algorithm uses a DropBlock regularization technique.

Similarly, to make up for an efficient object detection framework, the Mosaic data augmentation feature used in YOLO, allows mixing of 4 training images. This means that it helps mixing 4 different contexts, while CutMix being another data augmentation feature additionally mixes 2 input images. Due to this, YOLO is normally able to detect objects outside their normal context, which in our case being unique distresses within the pavement image. Since, batch normalization computes activation statistics from those 4 images on every layer, the requirement of a greater size for a mini-batch is significantly reduced. Also, while applying YOLO to the problem of predicting pavement distresses, proper selection of optimal hyper-parameters is necessary. In this study, the YOLO models are trained by setting the following hyperparameters: batch-size 64, the optimizer weight decay value of 0.0005, the initial learning rate of 0.01 and 0.937 momentum.

### B. CenterNet

CenterNet, just like YOLO is a state-of-the-art single stage object detection algorithm. However, it functions differently compared to YOLO or any other anchor based object detection methods. Technically, CenterNet is an advanced version of CornerNet [35]. The CornerNet is another anchor free method that uses a pair of corner key-points to localize objects. However, CenterNet unlike its predecessor, also uses the centered information, allowing it to use triplets rather than just a pair of keypoints. The network architecture of CenterNet is shown in Fig. 1. As seen from Figure 1, center pooling and cascade corner pooling is applied to enrich the center and corner information. Here, center pooling gathers useful and more identifiable visual patterns by extracting the maximum possible values in both horizontal and vertical directions. The method of center pooling follows the following order: first a feature map is obtained as an output from the backbone, then to verify if the pixel in a feature map is an actual center keypoint, max values from both its horizontal and vertical directions are figured out before adding them altogether. This, in turn helps center pooling obtain superior detection of center keypoints.

Similarly, cascade corner pooling collects the max values at both boundary condition and internal directions of an object. This feature helps in extracting not just the boundary information but also the objects' visual patterns. In the current task of detecting pavement distresses, CenterNet's Hourglass-104 backbone was used. While training CenterNet models, a learning rate of 0.000025 was used throughout the iteration. CenterNet had an average inference time of 340ms per image.

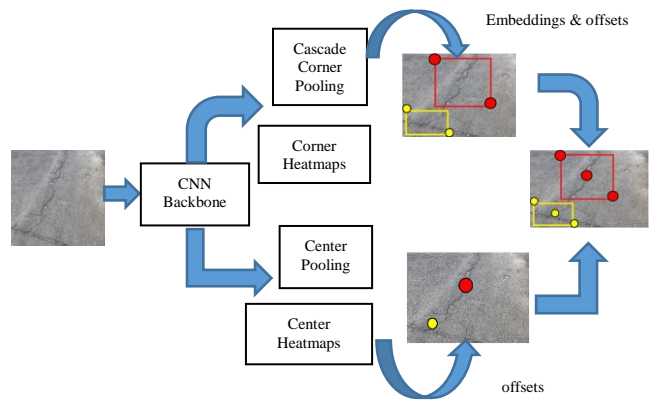

Fig. 1. Architecture of CenterNet

## C. EfficientDet

A scalable and efficient single-stage detector called EfficientDet [33] was deployed to predict distresses and compare the results of its performance with other detectors used in the study. EfficientDet combines EfficientNet [36] backbone with its bi-directional feature pyramid network (BiFPN) and the power of compound scaling. The BiFPN allows learnable weights to study the necessity of various input features and helps apply top-down and bottom-up fusion of features. Similarly, compound scaling helps scale the size of resolution, depth and width of the backbone, features, and the networks of bounding boxes and class prediction. In this study, EfficientDet model was trained using the SGD optimizer and momentum was set to 0.9 with the weight decay of 0.00004. Similarly, the learning rate was increased gradually from 0 to 0.16.

## D. Training and Post-Processing

With a view to speed up training and improve performance, transfer learning was used. In the field of computer vision and deep learning, transfer learning has been extensively used in optimizing learning techniques. Using this approach, the knowledge gained from the previous job could be seamlessly integrated to advance generalization about a new task. In this study, we used YOLO weights pre-trained on ImageNet as the initialization of the distress detection task. Also, multiple transfer learning strategies were analyzed using different weights. As training dataset consist of images from three different countries, the pavement and cracking textures were somewhat dissimilar from one another. Although, the images from Japan and Czech Republic, shared some similarity, they were still quite far-off from pavement images taken from India. Therefore, it is obvious that a single model trained on a comprehensive three country dataset would yield inferior detection and characterization results. So, taking heed of that, a model is trained to detect pavement distresses in Japan and Czech Republic images only, while a separate model is trained to detect distresses from India. While running on an NVIDIA GTX 1080 Ti GPU, the proposed models achieved an inference speed of approximately 65 frames per second.

Similarly, the size of YOLO based model proposed for detecting pavement distresses in Japan and Czech datasets, and trained for 54 epochs was roughly 381 megabytes whereas the model meant for India and trained for 100 epochs was about 361 megabytes. A heavier model aimed at Japan and Czech datasets and trained for 66 epochs averaged about 729 megabytes in size. Here, the total training time was around 15 hours. To understand the performance capability of YOLO based models, the network was trained for 300 epochs with a constant learning rate, weight decay and momentum. The models trained for different epochs were tested on the test1 and test2 datasets released by the IEEE Big Data, Global Road Damage Detection Challenge. Through these experimental results, it was observed that the models trained in between the range of 50-70 epochs performed comparatively better than those trained in between 150-300 epochs. Similarly, both CenterNet and EfficientDet were used for comparative analysis.

It was inferred that CenterNet and EfficientDet models trained for 300 epochs had similar performances to the ones trained for lower epochs. CenterNet and EfficientDet's overall training time for 300 epochs was roughly 75-80 hours separately on the same GPU hardware resources shared by YOLO.

In the post-processing stage, Intersection over Union (IoU) values were iterated and further analyzed for YOLO model. Although IOU is most commonly used as an evaluation metric, we also used it to remove duplicate bounding boxes, for the prediction of same pavement distresses. This was achieved by sorting all the predicted distresses present in an image in descending order of their confidence values. When two bounding boxes point out to the same distress type and class, its IOU is likely to have a very high value. In that case, the box with the highest IoU confidence would be chosen and the other one would be discarded.

## V. RESULTS

The performance of the proposed YOLO, CenterNet and EfficientDet models were evaluated on a set of 2,631 images present in test1 and 2,664 images in test2. In both test1 and test2, about 50% of the images were taken from Japan, 37% from India and around 13% from Czech Republic. F1 score, shown in equation (1) was used as an evaluation metric. The F1 score measures the model's accuracy using the harmonic mean of precision and recall. Here, precision as shown in equation (2) is the fraction of true positive (tp) examples amongst the retrieved samples i.e. true positives (tp) and false positives (fp). Similarly, recall shown in equation (3) is also known as sensitivity. It is the portion of the total number of relevant samples that were successfully retrieved. In other words, it is the ratio of true positives (tp) divided by true positives (tp) and false negatives (fn). For any distress prediction to be considered a true positive, it should have an Intersection over Union (IoU) value greater than or equal to 0.5 with the ground truth, and required both predicted and ground truth classes to have an exact match. Likewise, if a prediction obtained an IoU greater than or equal to 0.5 but classified a wrong distress class, then it was counted as false positive. The false negatives were referred to the ones that did not generate any predictions at all.

$$F1 = 2 \frac{precision \times \text{recall}}{precision + recall} \quad (1)$$

$$precision = \frac{tp}{tp + fp} \quad (2)$$

$$recall = \frac{tp}{tp + fn} \quad (3)$$

Some of the true positive examples of predicted pavement distresses for all three countries are shown in Fig. 2. The columns increasing from left to right show the longitudinal crack (D00), transverse crack (D10), alligator crack (D20) and potholes (D40). The predictions for Japan and Czech Republic

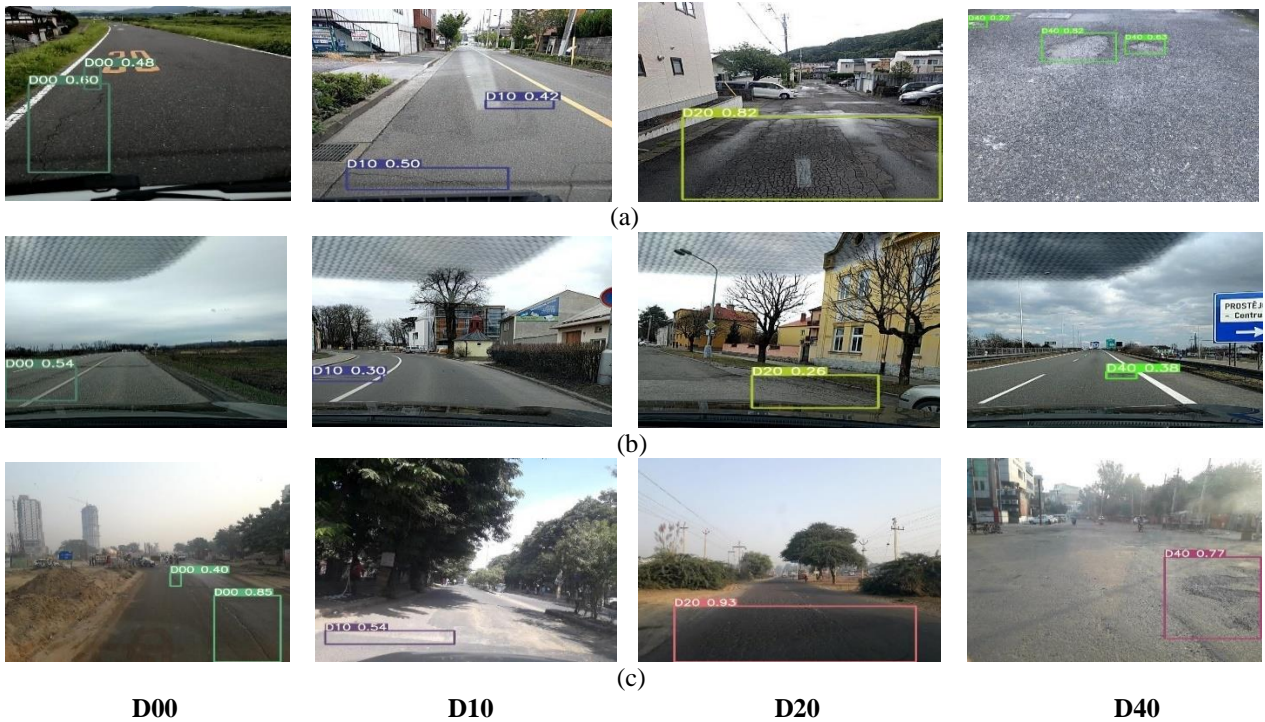

**D00**  **D10**  **D20**  **D40**

Fig. 2. Predicted Pavement Distresses: (a) Japan (b) Czech Republic (c) India

is made using the same YOLO model trained on 66 epochs whereas for India, a different model trained on 100 epochs is deployed. Similarly, some of the misclassifications or false positives are shown in Fig. 3 for all three algorithms. In Fig. 3(a) YOLO generates an incorrect prediction due to the presence of larger shadow on the upper portion of the image. It was observed that YOLO predicted erroneous detections for images with appreciable shadow formation. A possible resolve around this could be to augment the training database with sufficient number of images with all such scenarios. Similarly, in Fig. 3(b), CenterNet misclassified potholes and had several overlapping detections. Also, the bounding box size for alligator crack exceeds the regular cracking area. A similar issue is observed in Fig. 3(c) where the area of non-crack region exceeds the regular size and suffers from misclassification. EfficientDet suffers from the inability of detecting abstract features such as the one where the texture of distresses differs from the expected normal.

To attain the best prediction outcome, YOLO models were trained for different epochs and tested by varying the IoU and confidence thresholds. From our analysis, the models trained for 66 epochs on Japan and Czech dataset and 100 epochs on India dataset proved to be the most optimal. Similarly, for confidence thresholds lower than 0.1 would have several instances of false positives which would reduce precision. Also, increasing the IOU value greater than 0.9 would increase the possibility of predicting multiple overlapping boxes that could either potentially match or come closer to the ground truths. Experimental results demonstrate that the IoU of 1.0 and a confidence threshold of 0.17 would attain the best F1 score. Table II shows the test results obtained for all three models used in this study. YOLO achieved an F1 score of 0.5814 and 0.5751 on test1 and test2 respectively. It is interesting to note that both CenterNet and EfficientDet underperformed YOLO by almost 20 to 25 percent error margin, making YOLO a more suitable detector for this study. Although both CenterNet and

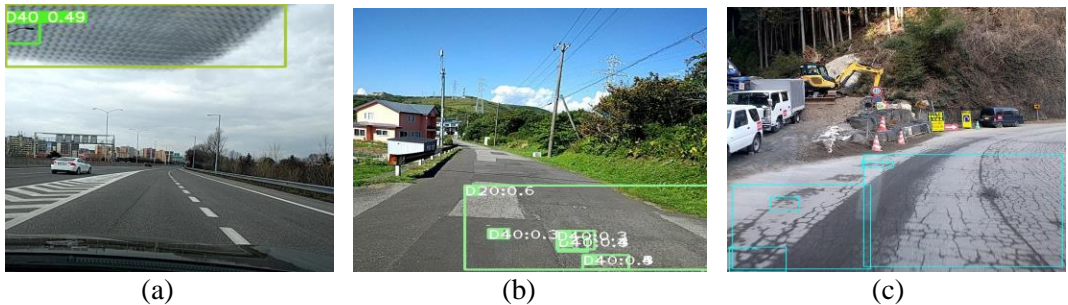

(a)  (b)  (c)

Fig 3. Distress Misclassification by Detectors: (a) YOLO (b) CenterNet (c) EfficientDet

EfficientDet possessed superior detection capability, they struggled at detecting transverse cracks and at times certain longitudinal cracks. While YOLO also faced, problems detecting longitudinal and transverse cracks, its performance at detecting such cracks on Japan images wasn't as substandard as it was on the other two. However, predicting distresses and characterizing them as transverse cracks appeared challenging. A deficit of transverse cracks in the training database could have been another possible reason. Similarly, despite having trouble detecting transverse cracks, both YOLO and CenterNet did well at predicting alligator cracks and potholes. EfficientDet, on the other hand faced difficulty at detecting alligator cracks and conceivably made predictions using larger bounding boxes covering greater portion of non-distresses.

TABLE II. Performance comparison of different algorithms

| Model | Backbone | Precision | Recall | F1 score |
| --- | --- | --- | --- | --- |
| test1 | | | | |
| YOLO | CSPDarknet53 | 0.59021 | 0.57296 | **0.5814** |
| CenterNet | Hourglass-104 | 0.48825 | 0.47665 | 0.4823 |
| EfficientDet | EfficientNet | 0.44269 | 0.43001 | 0.4362 |
| test2 | | | | |
| YOLO | CSPDarknet53 | 0.57986 | 0.57058 | **0.5751** |
| CenterNet | Hourglass-104 | 0.47865 | 0.47395 | 0.4762 |
| EfficientDet | EfficientNet | 0.44879 | 0.43958 | 0.4441 |

## VI. CONCLUSION

This study presents an automated approach to pavement distress detection and characterization using deep learning. Taking advantage of the pre-annotated database of training data, deep learning based models are trained on different network architectures. Post processing parameters such as the optimal confidence thresholds and increased IoU values were applied for different model versions. Overall, the best models achieved F1 scores of 0.5814 and 0.5751 on test1 and test2 datasets, released by the IEEE Global Road Damage Detection Challenge. Our observation suggest that these models performed generally well at detecting alligator cracks and potholes but had difficulty detecting transverse cracks. Especially, for the dataset available from India, very few instances of transverse cracks were seen. This caused a few misclassifications for images from India and prompted an unwarranted confusion with the longitudinal cracks. Also, the distress classes appeared somewhat unbalanced for the data available from India and Czech Republic. Therefore, some instances of false negative predictions were obtained. Regardless, a complete spectrum of distresses from Japan dataset helped the model achieve largely accurate predictions, which ultimately pushed the comprehensive three-country F1 score high. The study was conducted as part of the IEEE Global Road Damage Detection Challenge and the proposed solution was ranked 4[th] out of 121 teams worldwide, that participated in the challenge [37].

REFERENCES

[1] R. C. Detection. [Online]. Available: https://github.com/titanmu/RoadCrackDetection.
[2] T. I. USA Facts. [Online]. Available: https://usafacts.org/state-of-the-union/transportation-infrastructure/.
[3] A. Chatterjee and Y.-C. Tsai, "A fast and accurate automated pavement crack detection algorithm," in *2018 26th European Signal Processing Conference (EUSIPCO)*, 2018: IEEE, pp. 2140-2144.
[4] D. Arya *et al.*, "Transfer Learning-based Road Damage Detection for Multiple Countries," *arXiv preprint arXiv:2008.13101,* 2020.
[5] H. Maeda, Y. Sekimoto, T. Seto, T. Kashiyama, and H. Omata, "Road damage detection using deep neural networks with images captured through a smartphone," *arXiv preprint arXiv:1801.09454,* 2018.
[6] Y. J. Tsai and Z. Wang, "Development of an asphalt pavement raveling detection algorithm using emerging 3D laser technology and macrotexture analysis," 2015.
[7] K. C. Wang, Q. J. Li, G. Yang, Y. Zhan, and Y. Qiu, "Network level pavement evaluation with 1 mm 3D survey system," *Journal of traffic and transportation engineering (English edition),* vol. 2, no. 6, pp. 391-398, 2015.
[8] M. Gavilán *et al.*, "Adaptive road crack detection system by pavement classification," *Sensors,* vol. 11, no. 10, pp. 9628-9657, 2011.
[9] W. Vavrik, L. Evans, S. Sargand, and J. Stefanski, "PCR evaluation: considering transition from manual to semi-automated pavement distress collection and analysis," 2013.
[10] ASCE, "2017 infrastructure report card," 2017: ASCE Reston, VA.
[11] K. Gopalakrishnan, S. K. Khaitan, A. Choudhary, and A. Agrawal, "Deep convolutional neural networks with transfer learning for computer vision-based data-driven pavement distress detection," *Construction and Building Materials,* vol. 157, pp. 322-330, 2017.
[12] Y. Hu and C.-x. Zhao, "A novel LBP based methods for pavement crack detection," *Journal of pattern Recognition research,* vol. 5, no. 1, pp. 140-147, 2010.
[13] A. Miraliakbari, S. Sok, Y. O. Ouma, and M. Hahn, "Comparative Evaluation of Pavement Crack Detection Using Kernel-Based Techniques in Asphalt Road Surfaces," *International Archives of the Photogrammetry, Remote Sensing and Spatial Information Sciences,* vol. 1, 2016.
[14] M. Salman, S. Mathavan, K. Kamal, and M. Rahman, "Pavement crack detection using the Gabor filter," in *16th international IEEE conference on intelligent transportation systems (ITSC 2013)*, 2013: IEEE, pp. 2039-2044.
[15] T. Wang, K. Gopalakrishnan, A. K. Somani, O. G. Smadi, and H. Ceylan, "Machine-vision-based roadway health monitoring and assessment: Development of a shape-based pavement-crack-detection approach," 2016.
[16] Q. Zou, Y. Cao, Q. Li, Q. Mao, and S. Wang, "CrackTree: Automatic crack detection from pavement images," *Pattern Recognition Letters,* vol. 33, no. 3, pp. 227-238, 2012.
[17] W. Cao, Q. Liu, and Z. He, "Review of pavement defect detection methods," *IEEE Access,* vol. 8, pp. 14531-14544, 2020.
[18] L. Zhang, F. Yang, Y. D. Zhang, and Y. J. Zhu, "Road crack detection using deep convolutional neural network," in *2016 IEEE international conference on image processing (ICIP)*, 2016: IEEE, pp. 3708-3712.
[19] S. Li and X. Zhao, "Convolutional neural networks-based crack detection for real concrete surface," in *Sensors and Smart Structures Technologies for Civil, Mechanical, and Aerospace Systems, 2018*.
[20] B. Li, K. C. Wang, A. Zhang, E. Yang, and G. Wang, "Automatic classification of pavement crack using deep convolutional neural

network," *International Journal of Pavement Engineering,* vol. 21, no. 4, pp. 457-463, 2020.

[21] Z. Fan, Y. Wu, J. Lu, and W. Li, "Automatic pavement crack detection based on structured prediction with the convolutional neural network," *arXiv preprint arXiv:1802.02208,* 2018.

[22] M. D. Jenkins, T. A. Carr, M. I. Iglesias, T. Buggy, and G. Morison, "A deep convolutional neural network for semantic pixel-wise segmentation of road and pavement surface cracks," in *2018 26th European Signal Processing Conference (EUSIPCO)*, 2018: IEEE, pp. 2120-2124.

[23] Q. Zou, Z. Zhang, Q. Li, X. Qi, Q. Wang, and S. Wang, "Deepcrack: Learning hierarchical convolutional features for crack detection," *IEEE Transactions on Image Processing,* vol. 28, no. 3, pp. 1498-1512, 2018.

[24] W. Liu, Y. Huang, Y. Li, and Q. Chen, "FPCNet: Fast pavement crack detection network based on encoder-decoder architecture," *arXiv preprint arXiv:1907.02248,* 2019.

[25] J. Redmon and A. Farhadi, "YOLO9000: better, faster, stronger," in *Proceedings of the IEEE conference on computer vision and pattern recognition*, 2017, pp. 7263-7271.

[26] V. Mandal, L. Uong, and Y. Adu-Gyamfi, "Automated road crack detection using deep convolutional neural networks," in *2018 IEEE International Conference on Big Data (Big Data)*, 2018: IEEE, pp. 5212-5215.

[27] H. Majidifard, Y. Adu-Gyamfi, and W. G. Buttlar, "Deep machine learning approach to develop a new asphalt pavement condition index," *Construction and Building Materials,* vol. 247, p. 118513, 2020.

[28] J. Li, X. Zhao, and H. Li, "Method for detecting road pavement damage based on deep learning," in *Health Monitoring of Structural and Biological Systems XIII*, 2019, vol. 10972: International Society for Optics and Photonics, p. 109722D.

[29] S. Ren, K. He, R. Girshick, and J. Sun, "Faster r-cnn: Towards real-time object detection with region proposal networks," in *Advances in neural information processing systems*, 2015, pp. 91-99.

[30] H. Maeda, T. Kashiyama, Y. Sekimoto, T. Seto, and H. Omata, "Generative adversarial network for road damage detection," *Computer‐Aided Civil and Infrastructure Engineering*.

[31] A. Bochkovskiy, C.-Y. Wang, and H.-Y. M. Liao, "YOLOv4: Optimal Speed and Accuracy of Object Detection," *arXiv preprint arXiv:2004.10934,* 2020.

[32] K. Duan, S. Bai, L. Xie, H. Qi, Q. Huang, and Q. Tian, "Centernet: Keypoint triplets for object detection," in *Proceedings of the IEEE International Conference on Computer Vision*, 2019, pp. 6569-6578.

[33] M. Tan, R. Pang, and Q. V. Le, "Efficientdet: Scalable and efficient object detection," in *Proceedings of the IEEE/CVF Conference on Computer Vision and Pattern Recognition*, 2020, pp. 10781-10790.

[34] A. Paszke *et al.*, "Automatic differentiation in pytorch," 2017.

[35] H. Law and J. Deng, "Cornernet: Detecting objects as paired keypoints," in *Proceedings of the European Conference on Computer Vision (ECCV)*, 2018, pp. 734-750.

[36] M. Tan and Q. V. Le, "Efficientnet: Rethinking model scaling for convolutional neural networks," *arXiv preprint arXiv:1905.11946,* 2019.

[37] D. Arya *et al.*, "Global Road Damage Detection: State-of-the-art Solutions," *arXiv preprint arXiv: 2011.08740,* 2020.